\documentclass[runningheads]{llncs}

\usepackage{eccv}

\usepackage{eccvabbrv}

\usepackage{graphicx}
\usepackage{booktabs}
\usepackage{wrapfig}
\usepackage{array}
\usepackage{multirow}
\usepackage{makecell}
\usepackage{lipsum}
\usepackage{colortbl}
\usepackage{arydshln}
\usepackage{pifont}
\usepackage{xcolor}
\usepackage{soul}   
\usepackage{enumitem}

\newcommand{\hly}[1]{{\sethlcolor{yellow!75}\hl{#1}}}
\newcommand{\hlc}[1]{{\sethlcolor{violet!10}\hl{#1}}}
\usepackage{xcolor}
\usepackage{soul}
\sethlcolor{yellow} 

\definecolor{NS}{HTML}{FDC9C9} 
\definecolor{RS}{HTML}{C7EDFE} 
\definecolor{IE}{HTML}{B7E2B9} 
\definecolor{AN}{HTML}{E5BDF5} 

\newcommand{\hlyNS}[1]{{\sethlcolor{NS}\hl{#1}}}
\newcommand{\hlyRS}[1]{{\sethlcolor{RS}\hl{#1}}}
\newcommand{\hlyIE}[1]{{\sethlcolor{IE}\hl{#1}}}
\newcommand{\hlyAN}[1]{{\sethlcolor{AN}\hl{#1}}}

\definecolor{Gray}{gray}{0.85}
\definecolor{darkgreen}{rgb}{0.0, 0.5, 0.0} 
\definecolor{LightCyan}{rgb}{0.88,1,1}
\definecolor{LGray}{gray}{0.95}
\definecolor{lightred}{RGB}{255, 220, 220} 
\definecolor{mypurple}{RGB}{180,132,216}

\usepackage[accsupp]{axessibility}  
\makeatletter
\newcommand\blfootnote[1]{%
  \begingroup
  \renewcommand\thefootnote{}%
  \renewcommand\@makefnmark{}%
  \NoHyper
  \footnotetext{#1}%
  \endNoHyper
  \addtocounter{footnote}{-1}%
  \endgroup
}
\makeatother

\usepackage{hyperref}

\usepackage{orcidlink}

\begin{document}

\title{\textsc{C3-Bench}: A Context-Aware Change Captioning Benchmark} 


\author{
Jae-Woo Kim\inst{1}\orcidlink{0009-0007-1570-8174} \and
Hyeongbeom Kim\inst{1}\orcidlink{0009-0008-5147-2362} \and
Ue-Hwan Kim\inst{1,2}$^{\star}$\orcidlink{0000-0003-2201-2988}
}

\authorrunning{J. Kim et al.}

\institute{ \textsuperscript{1}Gwangju Institute of Science
and Technology, Gwangju, Republic of Korea\\  \textsuperscript{2}GIST InnoCORE AI-Nano Convergence Institute for Early Detection of Neurode-
generative Diseases, Gwangju Institute of Science and Technology, 61005 Gwangju, Republic of Korea \\ \email{kjw01124@gm.gist.ac.kr, hbk08101@gm.gist.ac.kr, uehwan@gist.ac.kr} \\[0.2em] \small\textbf{Project Page:} \texttt{\url{https://github.com/AutoCompSysLab/C3-Bench}}}

\maketitle
\blfootnote{* Corresponding author}

\begin{figure}
    \centering
    \vspace{-0.4cm}
    \includegraphics[width=0.9\linewidth]{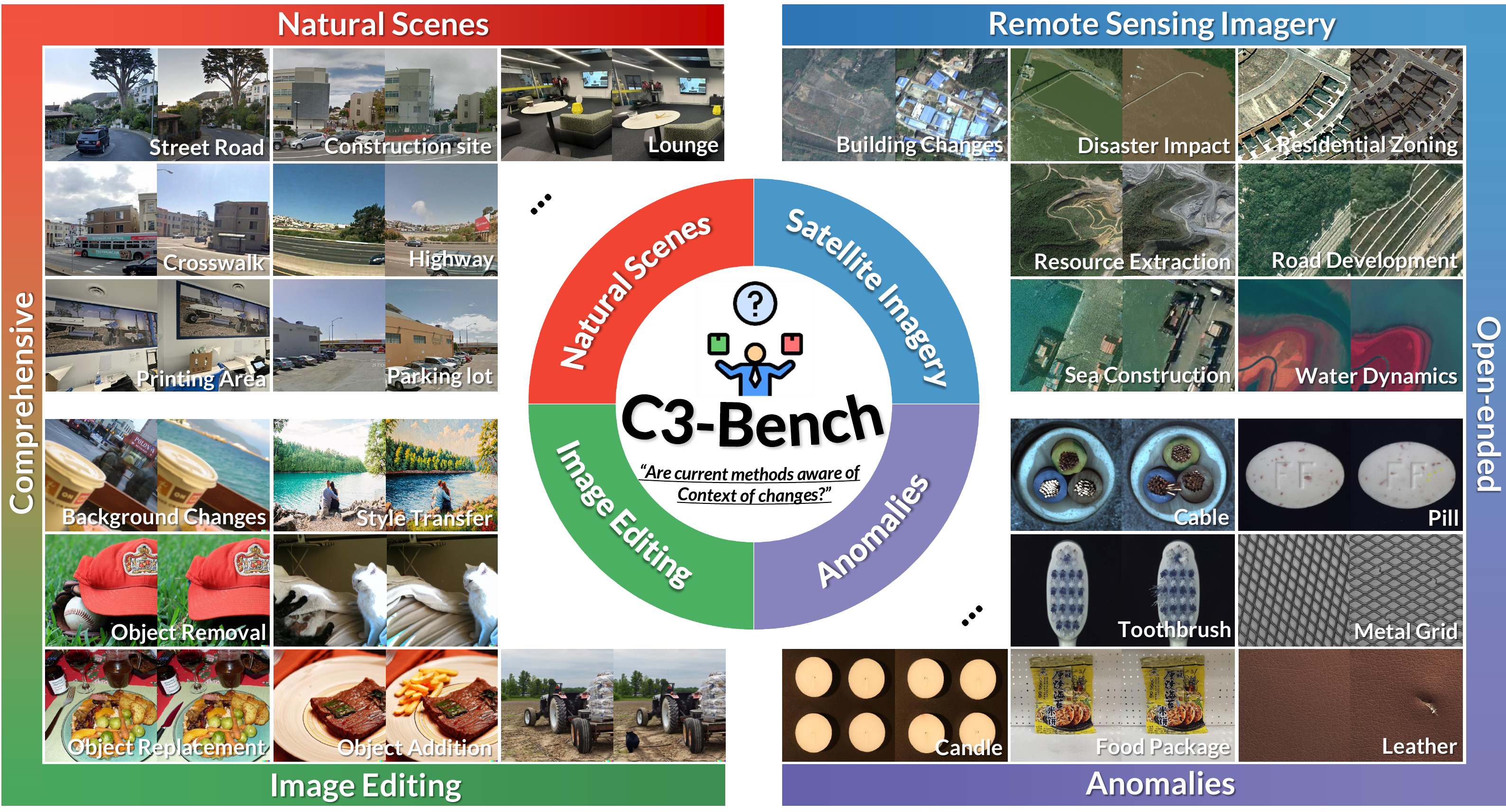}
    \caption{\textbf{Overview of \textsc{C3-Bench}}. The examples are from each context in \textsc{C3-Bench}.}
    \label{fig1}
    \vspace{-0.8cm}
\end{figure}

\begin{abstract}
While Change Captioning systems have garnered substantial attention to respond to our evolving world, their true performance on \textit{diverse real-world change contexts} remains largely unexplored due to the lack of comprehensive evaluation frameworks. To fill this gap, we propose \textbf{\textsc{C3-Bench}}, a comprehensive benchmark for evaluating \textbf{C}ontext-aware \textbf{C}hange \textbf{C}aptioning. \textsc{C3-Bench} features: (1) 4,996 human-labeled image pairs of 51 real-world change contexts across four domains (e.g., natural scenes, remote sensing imagery, image editing, and anomalies), each with diverse, carefully curated scenarios derived from multiple change-centric communities; and (2) the first LLM-as-Judge evaluation framework in the change captioning task that measure fine-grained dimensions (e.g., correctness, specificity, fluency, and relevance), along with a novel reversibility metric exploring whether models understand changes with symmetric consistency. Based on \textsc{C3-Bench}, we benchmark 32 models---including conventional change captioning models, proprietary Large Multimodal Models (LMMs), and 2B-90B open-source LMMs. We reveal a fundamental blind spot in the prevailing change captioning paradigm: Once the change context departs from training-style regimes, conventional models collapse, and even state-of-the-art LMMs such as GPT‑5.2 exhibit systematic domain- and position-dependent errors that distort reliable change understanding. By making these hidden failure modes explicit and measurable, we delineate the next frontier for building generalizable and trustworthy change captioning systems. All codes and datasets are publicly available on the project page.


\keywords{Change Captioning \and Dataset \& Benchmark}
\end{abstract}

\section{Introduction}
\label{sec:intro}

The world is non-stationary by default---structures emerge and decay \cite{Yang2022AsymmetricSN, Zheng2021ChangeIE}, objects appear and vanish \cite{Kim2021ViewpointAgnosticCC}, and environments drift under natural- and human-induced activities  \cite{Sakurada2017DenseOF, Gupta2019xBDAD}. Therefore, perceiving \textit{what has changed} and communicating \textit{how it changed} for intelligent systems to operate outside the controlled laboratories is no longer optional but \textit{indispensable} prerequisite for effective collaboration with humans; advancement here has an immediate impact on various downstream practices \cite{Du2014RemoteSI, Shafique2022DeepLC, Chen2023TimeTP, Chen2023ResolutionAgnosticRS, Zhang2024BiFARS, Chen2023ContinuousRS, Sun2024TheSD}---including incident monitoring, autonomy, anomaly reporting, remote sensing, and media forensics.

\begin{wrapfigure}{l}{6.4cm}
\vspace{-0.35cm}
\centering
\includegraphics[width=0.5\textwidth]{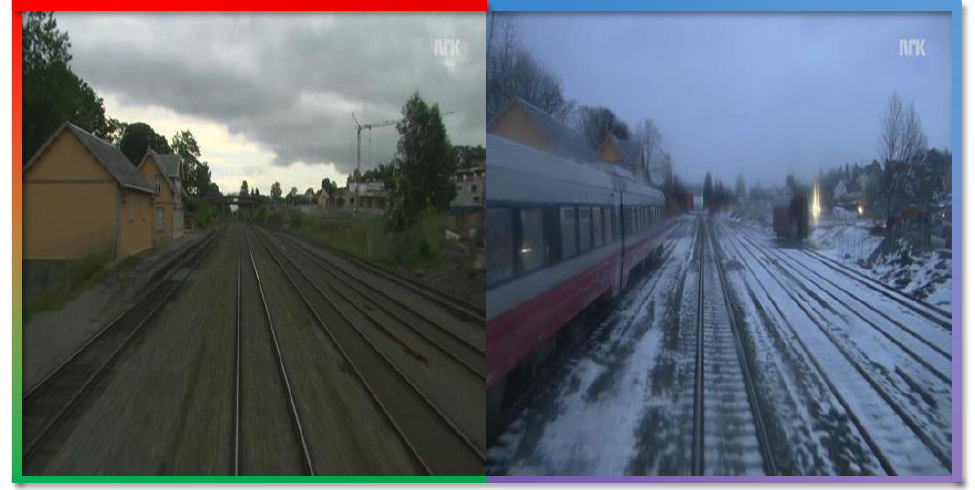}
    \caption{\textbf{What has changed? (Motivation)} Without a \textit{context}, this generic question can admit multiple logically valid descriptions.}
    \label{fig2}
\vspace{-0.45cm}
\end{wrapfigure}

Meanwhile, articulating change is fundamentally challenging due to the inherent ambiguity of ``\textit{change}''. Consider a single image pair shown in \cref{fig2}. When one is asked to describe the change between the given image pair, what might first come to mind is ``\textit{in which context?}'', as the definition of correct change can vary depending on the given \textbf{\textit{context}}: the valid description would be ``\textit{the snow has covered the ground, and cloud cover has decreased.}'' in respect of weather, whereas it is ``\textit{a train has appeared on the left side of the tracks.}'' for railway surveillance, with weather differences treated as pseudo-changes \cite{Alcantarilla2016StreetviewCD, Kim2024TowardsGS}. To meaningfully communicate and determine the correct change description among multiple logically valid alternatives in a heterogeneous visual world, each change must be grounded in \textit{specific contexts} and \textit{associated criteria} which clearly define the underlying semantics.

However, most prevailing benchmarks \cite{Jhamtani2018LearningTD, Tan2019ExpressingVR, Park2019RobustCC, Kim2021ViewpointAgnosticCC, Liu2022RemoteSI} largely overlook this point and describe changes either arbitrarily or implicitly---hindering a comprehensive understanding of diverse change contexts. Furthermore, their contextual coverage remains confined to narrow datasets---obscuring models' genuine performance in real-world scenarios. To bridge these gaps, we introduce \textbf{\textsc{C3-Bench}}---a comprehensive benchmark for evaluating \textbf{C}ontext-aware \textbf{C}hange \textbf{C}aptioning. Unlike previous works, \textsc{C3-Bench} offers a broader set of real-world change contexts (e.g, natural disaster, resource extraction, construction monitoring) with clearly-defined criteria, derived from multiple change-centric communities---such as change detection \cite{Alcantarilla2016StreetviewCD}, image editing \cite{Hui2024HQEditAH}, and anomaly detection \cite{Gao2025MetaUASUA}. As shown in \cref{fig1} and \cref{tab1}, \textsc{C3-Bench} consists of 4,996 image pairs spanning 51 real-world change contexts, organized into four overarching visual domains, each with diverse and carefully designed contexts within each domain. To collect data from disparate domains, we develop
an efficient annotation pipeline to produce high-quality
human-annotated data---reducing data contamination risks.

\begin{table*}[t]
\renewcommand{\arraystretch}{1.0}
\setlength{\aboverulesep}{0.5pt}
\setlength{\belowrulesep}{0.5pt}
\setlength\tabcolsep{0.8mm}
\setlength{\belowcaptionskip}{1.0mm}
\footnotesize
\centering
\caption{\textbf{Comparison with prior evaluation benchmarks}. Our \textsc{C3-Bench} covers a broad range of change-centric domains and contexts---providing a robust testbed for change captioning systems. See \cref{fig1} for detailed domains included in \textbf{ALL}.}

\resizebox{\linewidth}{!}{
\begin{tabular}{
    >{\raggedright\arraybackslash}p{4cm}
    >{\centering\arraybackslash}p{1.1cm}
    >{\centering\arraybackslash}p{1.7cm}
    >{\centering\arraybackslash}p{1.1cm}
    >{\centering\arraybackslash}p{2cm}
    >{\centering\arraybackslash}p{2cm}
    >{\centering\arraybackslash}p{2.2cm}
    >{\centering\arraybackslash}p{2cm}}
    \toprule
    \multirow{2.5}{*}{\textbf{Benchmark}} &  \multicolumn{3}{c}{\textbf{Dataset}} & \multicolumn{2}{c}{\textbf{Metrics}} & \multicolumn{2}{c}{\textbf{Baselines and Protocols}} \\
    \cmidrule(lr){2-4} \cmidrule(lr){5-6} \cmidrule(lr){7-8}
    & Domain & \# Contexts & \# Pairs & Open-ended & Reversibility & LMM-support & Open-domain \\
    \midrule
    \rowcolor{Gray} \textit{Synthetic Environment}  &&&&&&&\\ 
    CLEVR-Change \cite{Park2019RobustCC} & - & - & 7,950 & \textcolor{red}{\ding{55}} & \textcolor{red}{\ding{55}} & \textcolor{red}{\ding{55}} &\textcolor{red}{\ding{55}} \\

    CLEVR-DC \cite{Kim2021ViewpointAgnosticCC} & - & - & 4,800  & \textcolor{red}{\ding{55}} & \textcolor{red}{\ding{55}} & \textcolor{red}{\ding{55}} & \textcolor{red}{\ding{55}} \\
    \midrule
    \rowcolor{Gray} \textit{Real-world Environment}  &&&&&&&\\ 
    LEVIR-CC \cite{Liu2022RemoteSI} & Remote & 1 & 1,920 & \textcolor{red}{\ding{55}} & \textcolor{red}{\ding{55}} & \textcolor{red}{\ding{55}} & \textcolor{red}{\ding{55}} \\
    
    SpotTheDiff \cite{Jhamtani2018LearningTD} & Natural & 1 & 1,270 & \textcolor{red}{\ding{55}} & \textcolor{red}{\ding{55}} & \textcolor{red}{\ding{55}} & \textcolor{red}{\ding{55}} \\

    IER \cite{Tan2019ExpressingVR} & Editing & 3 & 495 & \textcolor{red}{\ding{55}} & \textcolor{red}{\ding{55}} & \textcolor{red}{\ding{55}} & \textcolor{red}{\ding{55}} \\

    
    
    \rowcolor{violet!10} \textbf{\textsc{C3-Bench} (Ours)} &  \textbf{ALL}  & \textbf{51} & \textbf{4,996} & \textcolor{darkgreen}{\checkmark} & \textcolor{darkgreen}{\checkmark} & \textcolor{darkgreen}{\checkmark} & \textcolor{darkgreen}{\checkmark} \\
    
    \bottomrule
\end{tabular}}
\label{tab1}
\end{table*}

In addition, many previous benchmarks rely on fixed reference-based scoring metrics such as BLEU \cite{Papineni2002BleuAM} and ROUGE \cite{Lin2004ROUGEAP}, which fail to capture the nuanced semantic difference of change descriptions. To overcome the challenges of evaluating open-ended change captioning, we establish LLM-as-Judge \cite{Gu2024ASO} metrics that score fine-grained dimensions of model predictions---facilitating a more human-aligned evaluation that reveals genuine progress toward real-world generalization. Moreover, we evaluate reversibility by swapping the input image order and examining whether the model understands changes in a symmetrically consistent manner---a crucial indicator for system reliability \cite{Kim2024TowardsGS}. We believe \textsc{C3-Bench} to lay a solid foundation for developing truly applicable change captioning systems.

Based on \textsc{C3-Bench}, we extensively benchmark 32 models---including conventional change captioning models, proprietary LMMs, and 2B-90B open-source LMMs. Our experiments yield several key findings: (1) Conventional models, despite their reported effectiveness, collapse in open-domain contexts---indicating brittle, dataset-tied notions of what count as change; and (2) While LMMs can close this gap under our criteria prompting, they remain limited by perceptual and spatial failures (Top-2 errors) and order-induced artifacts (pairwise position bias)---calling for more robust and reliable change captioning systems. Furthermore, our ablation study uncovers essential prompting principles for LMM-based change captioning, revealing that explicit conditioning on change criteria, sequential tags, and alignment with canonical temporal order are crucial.

To summarize, our contributions are threefold: 

\begin{enumerate}
    \item \textbf{Problem Formulation.} We define the task of Context-aware Change Captioning, a principled and delicate framework that consolidates heterogeneous change-centric problems into a single, unified task formulation.

    \item \textbf{Benchmark Construction.} We design \textsc{C3-Bench}, a comprehensive benchmark comprising 4,996 human-annotated image pairs spanning 51 real-world change contexts, alongside human-aligned evaluation metrics.
    \item \textbf{Critical Insights.} We conduct extensive experiments across diverse models, offering the first large-scale analysis of real-world performance and distilling critical insights into where and how research progress should be made.
\end{enumerate}

\section{Related Works}
\label{sec:related_works}

Most conventional methods have explored change captioning by enhancing model architectures \cite{Tan2019ExpressingVR, Shi2020FindingIA, Tu2021R3NetRelationembeddedRR, Qiu2021DescribingAL, Huang2022ImageDC, Guo2022CLIP4IDCCF, Yue2023I3NIA, Chang2023ChangesTC, Tu2023ViewpointAdaptiveRD, Tu2023NeighborhoodCT, Zhang2024DifferentialPerceptiveAR, Park2025LeveragingTC, Li2025RegionawareDD, Hu2025MCTCCDiffCC} and developing sophisticated training strategies \cite{Kim2021ViewpointAgnosticCC, Hosseinzadeh2021ImageCC, Yao2022ImageDC, Tu2023SelfsupervisedCR, Tu2024DistractorsImmuneRL, Lv2025RevisitingCC, Zhong2025DECIDERDC, anonymous2026imagine}. Their primary focus has been largely centered on improving performance within \textit{a single or a limited set of data-specific change contexts}, whereas the question of how to scale this task to \textit{a broader range of real-world change contexts} remains unanswered. Recently, LMMs such as GPT-4o \cite{Achiam2023GPT4TR} have shown promising results in understanding changes between a group of images \cite{Wang2024MuirBenchAC, Ying2024MMTBenchAC} or serving as pseudo-annotators \cite{Black2024VIXENVT, Hu2024OneDiffAG} to generate synthetic change captioning data. However, these efforts formulate change understanding as multiple-choice questions \cite{Fu2024BLINKML, Zhang2025AutomatedGO}, rather than requiring models to produce open-ended descriptions of visual changes. Moreover, they do not operate under clearly defined change criteria, relying on generic instructions such as ``\textit{describe the difference} \cite{Zhong2025DECIDERDC}.'' In contrast, our work focuses on open-ended change captioning grounded in a clearly defined context and criteria, enabling a more faithful and structured framework for change understanding.

Contemporary evaluation datasets such as CLEVR \cite{Johnson2016CLEVRAD}-based family \cite{Park2019RobustCC, Kim2021ViewpointAgnosticCC} primarily rely on synthetic tabletop scenes built from toy bricks---yielding overly simplified layouts with limited complexity. Real-world datasets \cite{Jhamtani2018LearningTD, Tan2019ExpressingVR, Liu2022RemoteSI} suffer from narrow topical and contextual coverage; LEVIR-CC \cite{Liu2022RemoteSI}, for example, largely comprises residential zoning over terrestrial surfaces---overlooking other crucial contexts such as natural disasters \cite{Gupta2019xBDAD, Zhang2023CrossdomainLM}, coastal expansion \cite{Shi2021ADS}, and land extraction \cite{Yu2024MineNetCDAB}. Furthermore, no existing dataset addresses multiple visual domains simultaneously, due to the context-agnostic problem formulation of existing approaches (see \cref{formulation}).

Evaluating change descriptions requires capturing nuanced semantic differences beyond surface-level lexical overlaps. However, most conventional metrics \cite{Sellam2020BLEURTLR, Banerjee2005METEORAA, Zhang2019BERTScoreET, Papineni2002BleuAM, Papineni2002BleuAM, Popovic2015chrFCN, Hessel2021CLIPScoreAR, Yuan2021BARTScoreEG, Anderson2016SPICESP, Lin2004ROUGEAP, Vedantam2014CIDErCI, Snover2006ASO, Han2012LEPORAR}, rely on fixed reference-based comparisons, failing to capture the semantic equivalence across diverse correct descriptions. This limitation becomes more pronounced in the case of LMMs, whose outputs are flexible, compositional, and often paraphrastic \cite{Yin2023ASO}. LLM-based evaluations have emerged as powerful alternatives \cite{Liu2023GEvalNE, Liu2024HolisticEF, Wu2024GPT4VisionIA, Vayani_2025_CVPR, Zhou2024OpenINGAC, Gu2024ASO}, however, their potential to change captioning remains largely underexplored. 

Lastly, although the change captioning task can now be performed by LMMs and is no longer exclusive to conventional models \cite{Hu2024OneDiffAG, Sun2024TheSD}, most evaluation frameworks still neither incorporate LMMs in a structured manner nor evaluate open-domain generalizability, focusing instead solely on in-domain performance within fixed datasets. To fill these gaps, we systematically evaluate and compare conventional change captioning models and LMMs in a unified framework, \textsc{C3-Bench}, revealing how models generalize across diverse real-world change contexts.
\section{Problem Formulation}
\label{formulation}

Given an image pair $\mathcal{I}_A$ and $\mathcal{I}_B$, conventional methods typically formulate change captioning as:
\begin{equation}
    \hat{y}_{A \rightarrow B} = \arg\max_{y \in \mathcal{Y}}
    p_{\theta}\bigl(y \mid \mathcal{I}_A, \mathcal{I}_B\bigr), \label{eq1}
\end{equation}
where $\hat{y}_{A \rightarrow B}$ denotes the generated change description from $\mathcal{I}_A$ to  $\mathcal{I}_B$, $\mathcal{Y}$ indicates the space of all textual descriptions, and $\theta$ is the model parameters. However, \cref{eq1} overlooks the context-dependent nature of changes. Moreover, it fails to capture the behavior of LMMs, which operate under multimodal conditioning \cite{Liu2023VisualIT, Chen2023LIONE, Wei2024RobustML} and incorporate textual instructions as part of their input. 

To account for this, we reformulate the \cref{eq1} as:
\begin{equation}
    \hat{y}^{c}_{A \rightarrow B} = \arg\max_{y \in \mathcal{Y}} p_{\theta}\bigl(y \mid { \mathcal{I}_A, \mathcal{I}_B }; \mathcal{C}_c\bigr), \label{eq2}
\end{equation}
where \(c\) indexes the context, \(\mathcal{C}_c\) denotes the associated criteria, and $\hat{y}^{c}_{A \rightarrow B}$ represents the model prediction that conforms to the semantics defined by $\mathcal{C}_c$. Here, \cref{eq1} can be considered as a special case of \cref{eq2} where $\mathcal{C}_c$ is unspecified. LMMs can be explicitly guided on $\mathcal{C}_c$ through textual instructions. When such specifications are absent---as in conventional models or generically prompted LMMs---the $\mathcal{C}_c$ becomes implicit (reduces to $\mathcal{C}_\theta$), reflecting the model's internal biases learned from the training distribution. Based on this formulation, we incorporate conventional models and instruction-guided LMMs within a unified change captioning framework, as well as establish a more comprehensive and controllable basis for change understanding.
\section{\textsc{C3-Bench}}

\subsection{Data Curation}
\label{data}

Collecting and labeling change captioning data is inherently challenging due to the costly nature of change annotation \cite{Zheng2021ChangeIE}. It is particularly difficult to gather and categorize images across diverse contexts while ensuring consistency \cite{Yang2024InvestigatingTE}. To this end, we created \textsc{C3-Bench} with an efficient pipeline described below.

\noindent\textbf{Context Conceptualization.} We begin by identifying diverse real-world phenomena where understanding change is crucial. To expand beyond conventional data scopes, we conduct an extensive literature review of diverse change-centric tasks across multiple communities, including change detection \cite{Alcantarilla2016StreetviewCD, Sakurada2017DenseOF, Sakurada2018WeaklySS, Varghese2018ChangeNetAD, Chen2021DRTANetDR}, image editing \cite{Tan2019ExpressingVR, Brooks2022InstructPix2PixLT, Zhang2023MagicBrushAM, Hui2024HQEditAH, Ryu2025TowardsSH}, and anomaly detection \cite{Bergmann2019MVTecA, Zou2022SPottheDifferenceSP, Zhang2023PKUGoodsADAS, Gao2025MetaUASUA}. By distilling the recurring patterns of change across these communities, we conceptualize 51 real-world change contexts, organized under four overarching domains.

\noindent\textbf{Context Formalization.} Building upon this context set, we formalize each context by defining context-specific change criteria $\mathcal{C}_c$ using a unified textual template (see \cref{tab:criteria}). Each set of criteria comprises three components: \textit{(i)} change to detect, specifying the target changes to be described; \textit{(ii)} change to ignore, outlining differences that are irrelevant or non-essential; and \textit{(iii)} tone and nuance, prescribing the desired linguistic style for expressing the detected changes \cite{Wang2023CaptionAI}. This process totals 51 context-criteria pairs that delineate the semantic boundaries of each change---facilitating systematic and consistent evaluation of change descriptions within a shared, well-defined semantic space.

    
\begin{table}[t]
\renewcommand{\arraystretch}{1.0}
\setlength{\aboverulesep}{0.5pt}
\setlength{\belowrulesep}{0.5pt}
\setlength\tabcolsep{0.8mm}
\setlength{\belowcaptionskip}{1.0mm}
\footnotesize
\centering
\caption{\textbf{Structure of context-specific change criteria.} The criteria consist of three parts: (\textit{i}) change to detect, (\textit{ii}) change to ignore, and (\textit{iii}) tone and nuance. The example below corresponds to the context, \textit{highway}, from the Natural Scenes domain.}

\resizebox{\linewidth}{!}{%
\begin{tabular}{
    >{\raggedright\arraybackslash}p{3cm}
    >{\raggedright\arraybackslash}p{11cm}}
    \toprule
    \textbf{Segment} & \multicolumn{1}{c}{\textbf{Description}} \\
    \midrule
    \multirow{5}{*}{Change to Detect} & $\bullet$  Movement, appearance, disappearance, or modification of vehicles, traffic signs, cones, or roadside objects. \\
    & $\bullet$  Structural or spatial rearrangements of lanes, barriers, or road markings. \\
    & $\bullet$ Functional or environmental state changes (e.g., traffic signs activated/deactivated). \\
    \hdashline
    \multirow{4}{*}{Change to Ignore}   & $\bullet$ Lighting or exposure variations \\
                                        & $\bullet$ Viewpoint or viewpoint-induced differences \\
                                        & $\bullet$ Minor photometric or compression artifacts\\
                                        & $\bullet$ Weather, seasonal, or shadow differences \\
    \hdashline
    Tone and Nuance & Neutral and objective.\\
    \bottomrule
\end{tabular}}
\label{tab:criteria}
\end{table}

\begin{figure}[t]
      \centering
      \includegraphics[width=\linewidth]{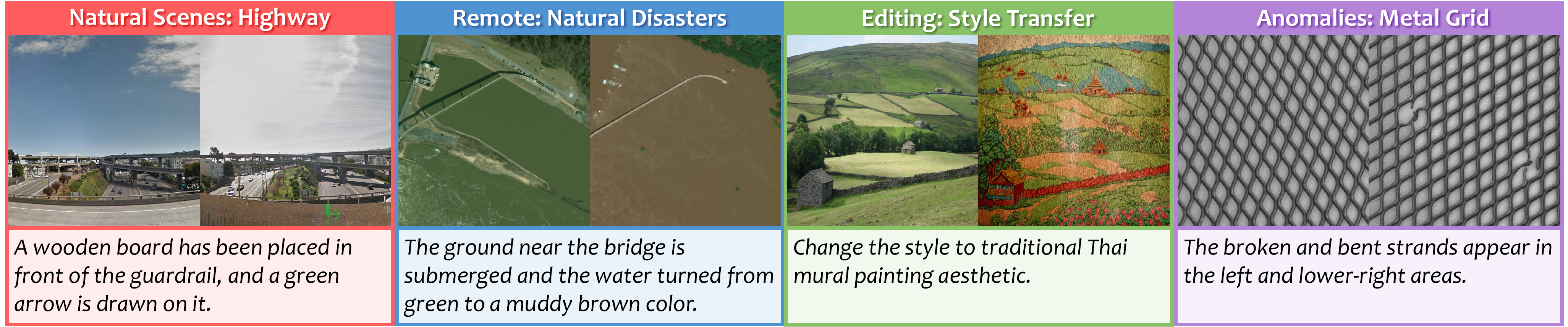}
    \caption{\textbf{Examples from \textsc{C3-Bench}.} Each image pair is displayed with its Domain: Context, along with the corresponding human-annotated change description.}
  \label{fig:sample}
\end{figure}

\noindent\textbf{Data Collection and Pairing.} We collect our change captioning data from more than fifteen publicly available benchmarks---including MVTecAD \cite{Bergmann2019MVTecA}, SECOND \cite{Yang2022AsymmetricSN}, MagicBrush \cite{Zhang2023MagicBrushAM}, PASLCD \cite{Galappaththige2024MultiViewPC}, and ChangeVPR \cite{Kim2024TowardsGS}. Unlike previous evaluation datasets, we incorporate heterogeneous and open-domain visual sources, substantially broadening data diversity \cite{Zhou2024OpenINGAC}. For crucial contexts not sufficiently covered by existing benchmarks, we supplement coverage by referring to publicly available imagery platforms such as Google Earth and Google Street View. We construct high-fidelity image pairs from existing correspondences or via feature matching \cite{Oquab2023DINOv2LR} and UTM-guided sampling \cite{Berton2022RethinkingVG, Kim2024TowardsGS}, followed by careful manual verification. Lastly, each pair is assigned to its corresponding context set and forwarded to the annotation stage.

\noindent\textbf{Data Annotation.} To ensure high-quality annotations, all annotators are provided with auxiliary metadata (e.g., change masks \cite{Zhang2023MagicBrushAM, Kim2024TowardsGS, Alcantarilla2016StreetviewCD}, building damage polygons \cite{Gupta2019xBDAD}, and anomaly categories \cite{Bergmann2019MVTecA, Zou2022SPottheDifferenceSP, Zhang2023PKUGoodsADAS}) overlaid on or displayed alongside the original images. Clear context-specific criteria are also provided to standardize the textual definition of relevant changes. All annotations are then converted into a unified schema for systematic downstream analysis.

\noindent\textbf{Data Filtering and Quality Control.} Each image pair is jointly reviewed by annotators and domain experts to ensure context-level consistency and semantic coherence. The candidate pairs then undergo a multi-stage validation pipeline---involving redundancy filtering, temporal verification, and caption-image consistency checks, followed by context-wise rebalancing to meet the target data size. \cref{fig:sample} illustrates several examples from the curated \textsc{C3-Bench}. A full list of contexts, data sources, auxiliary metadata, as well as further details and examples, is provided in the Appendix.

\subsection{Evaluation Metrics}

To facilitate more human-aligned evaluation, we adopt the LLM-as-Judge framework \cite{Zheng2023JudgingLW, Gu2024ASO}---to the best of our knowledge, for the first time in change captioning. We instruct GPT-5.2 \cite{openai_gpt5_2025} to assess each model-generated caption against the ground-truth description across four dimensions: correctness, specificity,  fluency, and relevance. The \textit{Correctness} metric refers to how closely the model prediction aligns with the ground-truth \cite{Liu2023VisualIT, Vayani_2025_CVPR}. The \textit{specificity} measures whether the predicted description provides a concrete explanation rather than a generic statement \cite{Hu2024AreLE}. The \textit{fluency} metric evaluates the tone and naturalness of model prediction \cite{Sai2020ASO, Wang2023CaptionAI}, whereas the \textit{relevance} metric determines whether model prediction provides answers directly related to the ground-truth \cite{Chakraborty2024TransferQS}. We also prompt GPT-5.2 to provide a single \textit{aggregation} score for a rapid, end-to-end comparison \cite{Zhou2024OpenINGAC}. To ensure fair evaluation of leveraging GPT-5.2 as a judge, we display consistent results using
Claude-Haiku-4.5 \cite{anthropic_claude_platform_api} in the Appendix.

Beyond these metrics, we also evaluate \textit{reversibility}, grounded in the premise that a reliable change understanding should remain semantically consistent under input-order reversal---differing only in directionality rather than exhibiting order-induced artifacts \cite{Kim2024TowardsGS, Tian2025IdentifyingAM}. Specifically, given a change description from $\mathcal{I}_A$ to $\mathcal{I}_B$ and its counterpart $\mathcal{I}_B$ to $\mathcal{I}_A$, we instruct GPT-5.2 to determine whether the two captions describe the same underlying changes with opposite directionality---assigning a score of 1 if reversible and 0 otherwise. This metric is partially related to position bias \cite{Tian2025IdentifyingAM} and temporal consistency \cite{Kim2024TowardsGS, wang2023reduce}, in that they all explore the robustness to input image ordering. However, fundamental differences are that we investigate reversibility at the level of open-ended descriptions and evaluate order effects on single image pairs rather than on long image sequences \cite{Wu2024VisualHA, Tian2025IdentifyingAM, Chen2024DocumentHV}. All evaluation prompts are provided in the Appendix. 
\section{Experiments and Results}
\label{experiments}

\subsection{Overall Setups}
We benchmark 32 models on \textsc{C3-Bench}---including 6 SOTA conventional change captioning models, 9 leading proprietary LMMs, and 17 open-source LMMs. The conventional models include DUDA \cite{Park2019RobustCC}, CLIP4IDC \cite{Guo2022CLIP4IDCCF}, VARD \cite{Tu2023ViewpointAdaptiveRD}, SCORER \cite{Tu2023SelfsupervisedCR}, and DIRL \cite{Tu2024DistractorsImmuneRL}. The proprietary LMMs involve GPT-5.2, GPT-5.1, GPT-5-mini, GPT-4o, and GPT-4o-mini  \cite{Achiam2023GPT4TR, openai_gpt5_2025}, Gemini-3-Pro, Gemini-3-Flash, Gemini-2.5-Pro, and Gemini-2.5-Flash \cite{google_aistudio_2025}. The open-source LMMs comprise Llama-3.2-Vision-Instruct (11B / 90B) and Llama-4.0-Scout-17B-16E \cite{Dubey2024TheL3}, Qwen2.5-VL (3B / 7B / 32B / 72B) \cite{Bai2025Qwen25VLTR} and Qwen3-VL (2B / 8B / 32B) \cite{Bai2025Qwen3VLTR}, InternVL3 (2B / 8B / 38B / 78B) \cite{zhu2025internvl3exploringadvancedtraining} and InternVL3.5 (2B / 8B / 38B) \cite{Wang2025InternVL35AO}. For consistency, all LMMs employ a temperature of 0 and a maximum completion length of 2048 tokens. Images are standardized to 512×512 pixel resolution. We adopt correctness, specificity, fluency, relevance, aggregation (all 1--10 scale), and reversibility rate (0--1) as our primary metrics. Since all changes are inherently reversible \cite{Kim2024TowardsGS}, the reversibility rate yields an upper bound of 1. 


\begin{table}[t]
\renewcommand{\arraystretch}{1.0}
\setlength{\aboverulesep}{0.5pt}
\setlength{\belowrulesep}{0.5pt}
\setlength\tabcolsep{0.8mm}
\setlength{\belowcaptionskip}{1.0mm}
\footnotesize
\centering
\caption{\textbf{System prompt for context-aware change captioning for LMMs}. The varying parts are marked in \hly{yellow}. During inference, these parts are dynamically replaced with the context-specific change criteria (\cref{tab:criteria}) and associated image pairs.}
\resizebox{\linewidth}{!}{%
\begin{tabular}{
    >{\raggedright\arraybackslash}p{3cm}
    >{\raggedright\arraybackslash}p{11cm}}
    \toprule
    \textbf{Segment} & \multicolumn{1}{c}{\textbf{Description}} \\
    \midrule
    \multirow{2}{*}{Task} & Analyze an image pair and describe the changes from the $<$before$>$ image to the $<$after$>$ image according to the given change criteria. \\
    \hdashline
    Change Criteria & \hly{$<$context-specific change criteria$>$} \\
    \hdashline
    \multirow{2}{*}{Output Format} & Produce a single coherent paragraph describing only the detected changes. Enclose the final answer within $<$output$>$ ... $<$$/$output$>$ tags.\\
    \hdashline
    \multirow{2}{*}{Image Pair} & Each image is preceded by a corresponding tag. \\
    & $<$before$>$ \hly{Image $\mathcal{I}_A$} $<$after$>$ \hly{Image $\mathcal{I}_B$}\\
    \bottomrule
\end{tabular}}

\label{tab:prompt}
\end{table}

\subsection{Evaluation Protocols}
\label{protocol}

\noindent\textbf{Humans.} We sample a subset of 400 image pairs---balanced across domains and contexts---which we refer to as \textsc{C3-Bench-small}. Human evaluators independently describe the changes in each image pair with the corresponding criteria, and their descriptions are scored using the aforementioned metrics. For comparison, we also report the performance of GPT‑5.2 on \textsc{C3-Bench-small}. Further details of the human evaluators' setups are provided in the Appendix.

\noindent\textbf{Conventional Models.} We train each conventional change captioning model on four representative datasets: CLEVR-CC \cite{Park2019RobustCC}, LEVIR-CC \cite{Liu2022RemoteSI}, SpotTheDiff \cite{Jhamtani2018LearningTD}, and IER \cite{Tan2019ExpressingVR}. This training process yields four distinctive models for each method. Then, we assess each of the four distinctive models on \textsc{C3-Bench}; this process yields 51 assessments per model, given the 51 context-criteria pairs. As \textsc{C3-Bench} subsumes a broad spectrum of real-world change contexts (including all real-world contexts covered by aforementioned datasets) with diverse data sources, the proposed protocol investigates the potential generalizability of the implicit semantics inherited in $\theta$ beyond data-specific performances.

\noindent\textbf{LMMs.} Depart from conventional models, LMMs can be explicitly conditioned on $\mathcal{C}_c$ through textual instructions. To facilitate fair and reproducible comparisons, we leverage all LMMs in an off-the-shelf setting, without any task-specific fine-tuning. We also avoid advanced prompt strategies \cite{Wei2022ChainOT, Wang2022SelfConsistencyIC} and multimodal prompting methods \cite{Guo2024RegionGPTTR, Baldassini2024WhatMM}. Instead, all LMMs are evaluated on \textsc{C3-Bench} using a structured prompt template shown in \cref{tab:prompt}. Here, a structured system prompt is combined with $\mathcal{C}_c$ in a plug-and-play manner for each of the 51 context-criteria pairs---explicitly capturing the contextual semantics. We also tag each input image pair to indicate its temporal order within the before--after sequence. During evaluation, the performance of \(\hat{y}^{c}_{A \rightarrow B}\) is compared to the ground-truth caption \(y^{c}_{A \rightarrow B}\) for each shared context $c$. Reversibility is evaluated by comparing \(\hat{y}^{c}_{A \rightarrow B}\) against \(\hat{y}^{c}_{B \rightarrow A}\) (for conventional models, \(\hat{y}_{A \rightarrow B}\) against \(\hat{y}_{B \rightarrow A}\)). This protocol parallels the context-wise evaluation used for conventional models, enabling unified comparison across model families while systematically investigating the capabilities of LMMs in a more principled and consistent manner.



\begin{table}[t]
\renewcommand{\arraystretch}{1.0}
\setlength{\tabcolsep}{1pt}
\footnotesize
\centering
\caption{\textbf{User study results.} We measure Pearson ($r$), Spearman ($\rho$), and Kendall ($\tau$) correlation coefficients to quantify alignment with human judgments.}
\label{tab:human}

\begin{minipage}[t]{0.5\columnwidth}
\centering
(a) Scoring-based judge. \par\vspace{0.2mm}
\setlength{\aboverulesep}{0.5pt}
\setlength{\belowrulesep}{0.5pt}

\resizebox{0.95\linewidth}{!}{
\begin{tabular}{
  >{\raggedright\arraybackslash}p{0.5\linewidth}
  >{\centering\arraybackslash}p{0.18\linewidth}
  >{\centering\arraybackslash}p{0.18\linewidth}
  >{\centering\arraybackslash}p{0.18\linewidth}}
\toprule
\textbf{Metric} & $r$ & $\rho$ & $\tau$ \\
\midrule
BLEU-4 \cite{Papineni2002BleuAM} & 0.3314 & 0.3202 & 0.2233 \\
ROUGE-L \cite{Lin2004ROUGEAP} & 0.4916 & 0.4460 & 0.3112 \\
chrF \cite{Popovic2015chrFCN} & 0.4548 & 0.4174 & 0.2879 \\
BERTScore \cite{Zhang2019BERTScoreET} & 0.4494 & 0.4443 & 0.3161 \\
CLIP-S \cite{Hessel2021CLIPScoreAR} & 0.2542 & 0.2931 & 0.2063 \\
\rowcolor{LGray} Aggregation & \textbf{0.7563} & \textbf{0.7944} & \textbf{0.6317} \\
\midrule
Correctness & 0.7173 & 0.7448 & 0.5963 \\
Specificity & 0.7268 & 0.7378 & 0.5812 \\
Fluency & 0.6193 & 0.5839 & 0.4857 \\
Relevance & 0.6382 & 0.6427 & 0.4985 \\
\bottomrule
\end{tabular}}
\end{minipage}\hfill
\begin{minipage}[t]{0.5\columnwidth}
\centering

(b) Pair-based judge.
\setlength{\aboverulesep}{0.5pt}
\setlength{\belowrulesep}{0.5pt}
\resizebox{0.95\linewidth}{!}{
\begin{tabular}{
  >{\raggedright\arraybackslash}p{0.5\linewidth}
  >{\centering\arraybackslash}p{0.18\linewidth}
  >{\centering\arraybackslash}p{0.18\linewidth}
  >{\centering\arraybackslash}p{0.18\linewidth}}
\toprule
\textbf{Metric} & $r$ & $\rho$ & $\tau$ \\
\midrule
BLEU-4 \cite{Papineni2002BleuAM} & 0.0146 & -0.0026 & -0.0021 \\
ROUGE-L \cite{Lin2004ROUGEAP} & 0.1129 & 0.1073 & 0.0880 \\
chrF \cite{Popovic2015chrFCN} & 0.1610 & 0.1596 & 0.1308 \\
BERTScore \cite{Zhang2019BERTScoreET} & 0.1017 & 0.0948 & 0.0777 \\
CLIP-S \cite{Hessel2021CLIPScoreAR} & -0.0107 & -0.0028 & -0.0023 \\
\rowcolor{LGray} Aggregation & \textbf{0.6118} & \textbf{0.6107} & \textbf{0.5156} \\
\bottomrule
\end{tabular}}
\end{minipage}

\vspace{-34pt}
\begin{minipage}[t]{0.5\columnwidth}
\centering
\end{minipage}\hfill
\makebox[\linewidth][r]{%
  \begin{minipage}[t]{0.5\columnwidth}
  \centering
  (c) Reversibility judge. \par\vspace{0.2mm}
  \setlength{\aboverulesep}{0.5pt}
  \setlength{\belowrulesep}{0.5pt}
  \resizebox{0.9\linewidth}{!}{
    \begin{tabular}{
      >{\raggedright\arraybackslash}p{0.4\linewidth}
      >{\centering\arraybackslash}p{0.19\linewidth}
      >{\centering\arraybackslash}p{0.19\linewidth}
      >{\centering\arraybackslash}p{0.19\linewidth}}
    \toprule
    \textbf{Metric} & $r$ & $\rho$ & $\tau$ \\
    \midrule
    \rowcolor{LGray} Reversibility & \textbf{0.8674} & \textbf{0.7670} & \textbf{0.7025} \\
    \bottomrule
    \end{tabular}}
  \end{minipage}%
}
\label{tab:user_study}
\end{table}


\subsection{Alignment with Human Judgments}

Before benchmarking performance on \textsc{C3-Bench}, we validate our proposed evaluation metrics with human judgments through a user study involving 40 participants on \textsc{C3-Bench-small} (see \cref{tab:human}). We sample 2,800 change descriptions generated by seven LMMs (the latest models from each generation), and split them into 1,400 instances for scoring-based and 1,400 for pair-based protocols \cite{Chen2024MLLMasaJudgeAM}; additional details on the user study are provided in the Appendix. The results show that all proposed metrics exhibit consistently stronger alignment with human judgments than conventional metrics across both protocols. Surprisingly, GPT-5.2 achieves a strong correlation with human judgments on reversibility (Pearson $r$ > 0.80)---indicating high fidelity as a reversibility judge and consistent with prior observations that large language models can approximate human judgments \cite{Dubois2023AlpacaFarmAS, Zheng2023JudgingLW, Fu2023GPTScoreEA, Zhang2023GPT4VisionAA, Liu2023GEvalNE, Chen2024MLLMasaJudgeAM}. Importantly, beyond improved human alignment, our metrics uniquely enable fine-grained analysis of individual change descriptions while remaining fully compatible with rapid, end-to-end comparisons. Collectively, these contributions establish a human-aligned foundation within our \textsc{C3-Bench} framework that reveals genuine yet previously obscured real-world generalization of contemporary change captioning approaches.

\begin{figure}[t]
    \centering
    \includegraphics[width=\linewidth]{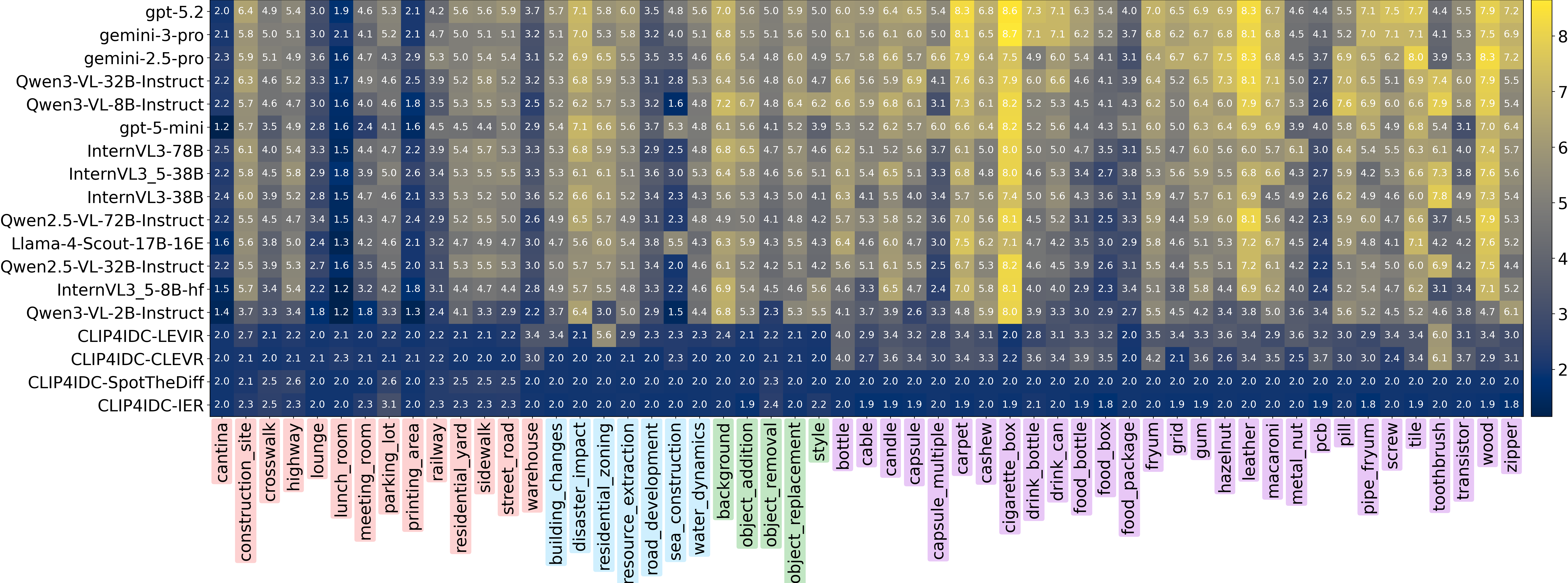}
    \vspace{-18pt}
    \caption{\textbf{Fine-grained results across domains and contexts on \textsc{C3-Bench}.} The performance is represented as an Aggregation score, where each context is highlighted with its corresponding domain: \hlyNS{Natural Scenes}, \hlyRS{Remote Sensing}, \hlyIE{Image Editing}, and \hlyAN{Anomalies}. The actual score of each model on a given context is shown in the corresponding cell, with higher scores indicated by higher color intensity. } 
    \label{fig:fine}
\end{figure}

\begin{table*}[t!]
\renewcommand{\arraystretch}{1.0}
\setlength{\aboverulesep}{0.5pt}
\setlength{\belowrulesep}{0.5pt}
\setlength\tabcolsep{0.8mm}
\setlength{\belowcaptionskip}{1.0mm}
\footnotesize
\centering
\caption{\textbf{\textsc{C3-Bench} results}. Mean and standard deviation are reported over three GPT-5.2 runs. $^{\dagger}$ denotes results on \textsc{C3-Bench-small}. Conventional models are averaged over trained weights. Human scores are in \hlc{violet}; best model scores are in \textbf{bold}.}
\resizebox{\linewidth}{!}{%
\begin{tabular}{l
    >{\centering\arraybackslash}p{1.8cm}
    >{\centering\arraybackslash}p{1.8cm}
    >{\centering\arraybackslash}p{1.8cm}
    >{\centering\arraybackslash}p{1.8cm}
    >{\centering\arraybackslash}p{1.8cm}
    >{\centering\arraybackslash}p{1.8cm}}
    \toprule
    \textbf{Method} &
    Correctness &
    Specificity &
    Fluency &
    Relevance &
    Aggregation &
    Reversibility \\
    \midrule
    \rowcolor{Gray} \textit{Conventional Models}  &&&&&&\\ 
    DUDA \cite{Park2019RobustCC} & $1.39\pm0.01$ &  $1.43\pm0.05$ & $7.81\pm0.25$ & $2.11\pm0.23$ & $2.09\pm0.09$ & $0.15\pm0.00$ \\

    $\text{R}^3$Net \cite{Tu2021R3NetRelationembeddedRR} & $1.39\pm0.01$ &  $1.61\pm0.09$ & $8.26\pm0.11$ & $2.67\pm0.19$ & $2.39\pm0.06$ & $0.13\pm0.00$ \\

    CLIP4IDC \cite{Guo2022CLIP4IDCCF} & $1.40\pm0.02$ &  $1.78\pm0.05$ & $8.07\pm0.13$ & $2.77\pm0.20$ & $2.43\pm0.06$ & $0.21\pm0.00$ \\

    SCORER \cite{Tu2023SelfsupervisedCR} & $1.39\pm0.01$ &  $1.56\pm0.07$ & $7.98\pm0.12$ & $2.74\pm0.20$ & $2.36\pm0.06$ & $0.18\pm0.00$ \\

    VARD \cite{Tu2023ViewpointAdaptiveRD} & $1.44\pm0.02$ &  $1.56\pm0.06$ & $8.29\pm0.12$ & $2.75\pm0.20$ & $2.40\pm0.07$ & $0.28\pm0.00$ \\
    
    DIRL \cite{Tu2024DistractorsImmuneRL} & $1.36\pm0.01$ &  $1.53\pm0.05$ & $7.73\pm0.15$ & $2.71\pm0.22$ & $2.30\pm0.04$ & $0.20\pm0.00$ \\
    \midrule
    \rowcolor{Gray} \textit{\textsc{C3-Bench-small} Performance}  &&&&&&\\ 

    ${\text{Human Level}}^{\dagger}$ & \cellcolor{violet!10} $6.96\pm0.10$ & \cellcolor{violet!10} $7.12\pm0.08$ & \cellcolor{violet!10} $8.32\pm0.23$ & \cellcolor{violet!10} $7.89\pm0.15$ & \cellcolor{violet!10} $7.45\pm0.18$ & \cellcolor{violet!10} $0.93\pm0.00$ \\
    
    ${\text{GPT-5.2}}^{\dagger}$ \cite{openai_gpt5_2025}& $5.64\pm0.12$ &  $5.27\pm0.17$ &$8.53\pm0.25$ & $6.33\pm0.17$ & $5.72\pm0.20$ & $0.61\pm0.00$ \\
    
    \midrule
    \rowcolor{Gray} \textit{Proprietary LMMs}  &&&&&&\\ 
    
    GPT-4o-mini \cite{Achiam2023GPT4TR} & $4.56\pm0.09$ &  $4.16\pm0.13$ &$8.67\pm0.23$ & $5.23\pm0.16$ & $4.65\pm0.17$ & $0.29\pm0.00$ \\ 
    
    GPT-4o \cite{Achiam2023GPT4TR} & $4.95\pm0.10$ &  $4.34\pm0.15$ &$8.48\pm0.36$ & $5.83\pm0.26$ & $4.95\pm0.19$ & $0.49\pm0.00$ \\

    GPT-5-mini \cite{openai_gpt5_2025}& $4.87\pm0.13$ &  $4.60\pm0.21$ &$8.34\pm0.22$ & $5.00\pm0.16$ & $4.87\pm0.19$ & $0.45\pm0.00$ \\ 

    GPT-5.1 \cite{openai_gpt5_2025} & $5.43\pm0.13$ &  $5.12\pm0.20$ &$8.61\pm0.29$ & $5.98\pm0.17$ & $5.49\pm0.20$ & $0.63\pm0.00$ \\

    GPT-5.2 \cite{openai_gpt5_2025}& $\mathbf{5.47\pm0.11}$ &  $\mathbf{5.16\pm0.18}$ &$8.65\pm0.30$ & $\mathbf{5.98\pm0.13}$ & $\mathbf{5.51\pm0.20}$ & $0.62\pm0.00$ \\
    
    \midrule
    
    Gemini-2.5-Flash \cite{google_aistudio_2025} & $4.99\pm0.13$ &  $4.69\pm0.17$ &$8.26\pm0.33$ & $5.69\pm0.15$ & $5.04\pm0.20$ & $0.48\pm0.00$ \\
    
    Gemini-2.5-Pro \cite{google_aistudio_2025} & $5.18\pm0.10$ &  $4.92\pm0.15$ &$8.37\pm0.32$ & $5.85\pm0.12$ & $5.23\pm0.18$ & $0.52\pm0.00$ \\

    Gemini-3-Flash \cite{google_aistudio_2025} & $5.42\pm0.10$ &  $5.08\pm0.14$ &$8.34\pm0.30$ & $5.86\pm0.11$ & $5.39\pm0.17$ & $0.62\pm0.00$ \\
    
    Gemini-3-Pro \cite{google_aistudio_2025} & $5.45\pm0.10$ &  $5.15\pm0.16$ &$8.40\pm0.25$ & $5.95\pm0.17$ & $5.50\pm0.19$ & $\mathbf{0.73\pm0.00}$ \\
    \midrule
    
    \rowcolor{Gray} \textit{Open-source LMMs}  &&&&&&\\ 
    
    Llama-3.2-11B-Vision-Ins. \cite{Dubey2024TheL3} & $2.45\pm0.06$ &  $2.51\pm0.12$ &$8.14\pm0.29$ & $2.89\pm0.13$ & $2.56\pm0.12$ & $0.12\pm0.00$ \\
    
    Llama-3.2-90B-Vision-Ins. \cite{Dubey2024TheL3} & $2.55\pm0.10$ &  $2.62\pm0.13$ &$8.20\pm0.20$ & $2.93\pm0.11$ & $2.65\pm0.14$ & $0.20\pm0.00$ \\

    Llama-4-Scout-17B-16E-Ins. \cite{Dubey2024TheL3} & $4.63\pm0.09$ &  $4.43\pm0.13$ &$8.46\pm0.29$ & $5.19\pm0.23$ & $4.71\pm0.17$ & $0.44\pm0.00$ \\

    \midrule
    
    Qwen2.5-VL-3B-Ins. \cite{Bai2025Qwen25VLTR} & $2.87\pm0.04$ &  $2.75\pm0.07$ &$8.21\pm0.21$ & $3.90\pm0.21$ & $3.00\pm0.15  $ & $0.60\pm0.00$ \\
    
    Qwen2.5-VL-7B-Ins. \cite{Bai2025Qwen25VLTR} & $4.26\pm0.09$ &  $3.99\pm0.14$ &$8.63\pm0.27$ & $5.07\pm0.21$ & $4.41\pm0.18$ & $0.31\pm0.00$ \\
    
    Qwen2.5-VL-32B-Ins. \cite{Bai2025Qwen25VLTR} & $4.51\pm0.11$ &  $4.28\pm0.16$ &$8.30\pm0.23$ & $5.06\pm0.25$ & $4.57\pm0.20$ & $0.40\pm0.00$ \\
    
    Qwen2.5-VL-72B-Ins. \cite{Bai2025Qwen25VLTR} & $4.54\pm0.11$ &  $4.38\pm0.17$ &$8.36\pm0.29$ & $5.06\pm0.23$ & $4.61\pm0.19$ & $0.38\pm0.00$ \\

    \midrule
    
    Qwen3-VL-2B-Ins. \cite{Bai2025Qwen3VLTR}  & $3.84\pm0.10$ &  $3.74\pm0.16$ &$8.56\pm0.33$ & $4.38\pm0.24$ & $3.92\pm0.19$ & $0.28\pm0.00$ \\

    Qwen3-VL-8B-Ins. \cite{Bai2025Qwen3VLTR}  & $5.07\pm0.12$ &  $4.76\pm0.18$ &$8.68\pm0.32$ & $5.54\pm0.13$ & $5.10\pm0.21$ & $0.40\pm0.00$ \\

    Qwen3-VL-32B-Ins. \cite{Bai2025Qwen3VLTR}  & $5.18\pm0.14$ &  $4.93\pm0.22$ &$8.49\pm0.32$ & $5.58\pm0.18$ & $5.16\pm0.23$ & $0.47\pm0.00$ \\

    \midrule
    
    InternVL3-2B \cite{zhu2025internvl3exploringadvancedtraining} & $3.19\pm0.06$ &  $2.96\pm0.10$ &$8.72\pm0.21$ & $4.08\pm0.26$ & $3.35\pm0.13$ & $0.28\pm0.00$ \\

    InternVL3-8B \cite{zhu2025internvl3exploringadvancedtraining}& $3.93\pm0.07$ &  $3.68\pm0.10$ &$8.78\pm0.22$ & $4.77\pm0.22$ & $4.10\pm0.16$ & $0.36\pm0.00$ \\

    InternVL3-38B \cite{zhu2025internvl3exploringadvancedtraining}& $4.72\pm0.09$ &  $4.34\pm0.13$ &$8.16\pm0.19$ & $5.46\pm0.29$ & $4.72\pm0.19$ & $0.46\pm0.00$ \\

    InternVL3-78B \cite{zhu2025internvl3exploringadvancedtraining} & $4.86\pm0.09$ &  $4.51\pm0.14$ &$8.66\pm0.28$ & $5.59\pm0.24$ & $4.94\pm0.19$ & $0.45\pm0.00$ \\
    
    \midrule
     
    InternVL3.5-2B \cite{Wang2025InternVL35AO} & $3.20\pm0.06$ &  $3.08\pm0.10$ &$\mathbf{8.94\pm0.23}$ & $4.32\pm0.20$ & $3.41\pm0.18$ & $0.44\pm0.00$ \\

    InternVL3.5-8B \cite{Wang2025InternVL35AO}& $4.30\pm0.08$ &  $4.16\pm0.13$ &$8.63\pm0.23$ & $4.92\pm0.21$ & $4.44\pm0.16$ & $0.33\pm0.00$ \\

    InternVL3.5-38B \cite{Wang2025InternVL35AO} & $4.83\pm0.08$ &  $4.52\pm0.13$ &$8.46\pm0.28$ & $5.54\pm0.24$ & $4.92\pm0.18$ & $0.48\pm0.00$ \\
    
    \bottomrule
 \end{tabular}}
\label{tab:performance1}
\end{table*}

\subsection{Benchmark Results}

\cref{tab:performance1} summarizes the results of 32 models on \textsc{C3-Bench}. We illustrate our key observations as follows:

\noindent\textbf{Human Level Performance.} Not surprisingly, human evaluators demonstrate high performance across all semantic dimensions---outperforming the best LMM (GPT-5.2) by 1.73 points in Aggregation. Furthermore, human performance on the reversibility rate is remarkably high (0.93)---indicating a strong intuitiveness in understanding symmetrical changes. In contrast, the gap narrows considerably on Fluency, suggesting LMM-generated sentences can be comparable or even superior in linguistic coherence and naturalness---demonstrating their effectiveness in producing high-quality textual content \cite{Zhou2024OpenINGAC}.

\noindent\textbf{Limited Generalization of Conventional Models.}
Across \textsc{C3-Bench}, conventional methods exhibit consistently low performance, while maintaining relatively high Fluency metric (7.73--8.29). Notably, this performance collapse persists even for recent state-of-the-art models such as DIRL \cite{Tu2024DistractorsImmuneRL}, which have reported strong results on previous evaluation benchmarks. These findings underscore two broader implications: \textit{(i)} while conventional models can generate fluent descriptions, they fail to capture the intended change semantics under diverse real-world change contexts; and \textit{(ii)} existing evaluation benchmarks, largely optimized for fixed reference matching with limited contextual diversity, may not stress-test a model’s ability to adapt to diverse real-world changes. 

\noindent\textbf{LMM Landscapes in Change Captioning.} In general, proprietary models achieve higher Aggregation scores, ranging from 4.65 to 5.51, led by GPT-5.2. For open-source models, small variants (e.g., Llama-3.2-Vision-11B or Qwen2.5-VL-3B) surpass conventional models but remain limited in semantic understanding. Qwen3-VL-32B delivers strong overall performance, yielding an Aggregation of $5.16 \pm 0.23$ that matches Gemini-2.5-Pro ($5.23 \pm 0.18$) and trails GPT-5.2 by only 0.35 points. These findings indicate that open-source LMMs have almost bridged the gap with proprietary models, with remaining challenges mainly arising from semantic accuracy rather than linguistic coherence. Overall, the results exhibit a monotonic gain as LMMs scale in size and advance across generations, consistent with observations reported in recent multimodal benchmarks \cite{Yu2023MMVetEL, Yue2024MMMUProAM, Ying2024MMTBenchAC, Chen2024AreWO}.

\noindent\textbf{Fine-grained Results across Domains and Contexts.} As shown in \cref{fig:fine}, several models, including GPT-5-mini and Llama-4-Scout-17B-16E, exhibit pronounced strengths on specific contexts such as sea constructions \cite{Shi2021ADS} (in Remote Sensing), indicating that their gains are often highly localized rather than uniformly distributed across contexts. At the domain level, open-source models such as Qwen3-VL-8B surpass proprietary models in the Image Editing domain. Conventional models peak on only a few contexts and perform poorly elsewhere, due to architectural constraints that limit them to fixed, training-time semantics; further discussion on this and directions for future work are in the Appendix.

\noindent\textbf{Do Models Understand Change Symmetrically?} Notably, we discover that conventional models exhibit limited reversibility, revealing severe order sensitivity and a failure to form stable, order-invariant change representations. While LMMs show improved robustness, significant order bias remains: Gemini-3-Pro achieves the highest reversibility of 0.73; yet, open-source LMMs exhibit stronger sensitivity to image ordering---hindering a reliable understanding of changes \cite{Kim2024TowardsGS}. This highlights that position bias \cite{Tian2025IdentifyingAM, Wu2024VisualHA} in LMMs is not confined to long image sequences (e.g., 5–100 images) but is already pronounced in simple image pairs.

\begin{figure}[t]
    \centering
    \includegraphics[width=0.9\linewidth]{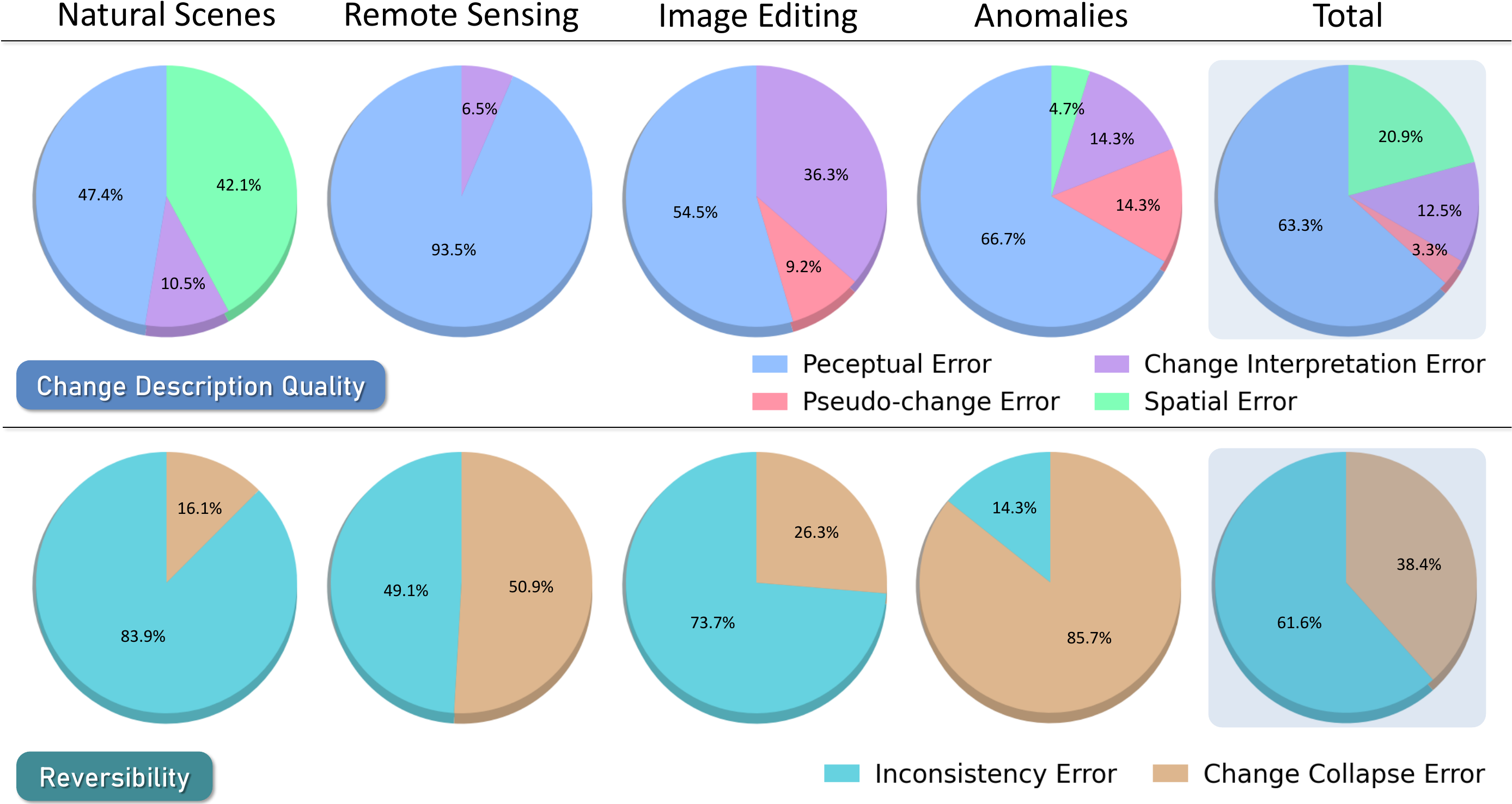}
    \vspace{-2pt}
    \caption{\textbf{Error distributions across domains.} \textit{(i)} change description quality errors (Top); and \textit{(ii)} reversibility errors (Bottom).}
    \label{fig:error1}
\end{figure}

\subsection{Error Analysis} 
\label{analysis}
To quantify and characterize the main bottlenecks of the best-performing model on our benchmark, we investigate its errors on \textsc{C3-Bench-small}. We first categorize failures related to change description quality into four distinct types and provide a clear definition upon examination:
\begin{enumerate}[noitemsep, topsep=0pt]
    \item \textbf{Perceptual Error}, arising from unrecognized objects, misclassified object categories, or incorrect object attributes.
    \item \textbf{Pseudo-change Error}, describing context-irrelevant objects or states as meaningful changes.
    \item \textbf{Change Interpretation Error}, failing to correctly infer how it has changed between the given image pair.
    \item \textbf{Spatial Error}, misinterpreting the spatial relationships or locations associated with the changes.
\end{enumerate}
\noindent Further, we define two reversibility-related errors as follows:
\begin{enumerate}[noitemsep, topsep=0pt]
    \item\textbf{Inconsistency Error}, producing semantically inconsistent descriptions that are not mutually inverse with omitted or extraneous key details.
    \item\textbf{Change Collapse Error}, identifying a change in one direction but failing to detect a change at all in the reversed order.
\end{enumerate}

\begin{table*}[t]
\renewcommand{\arraystretch}{1.0}
\setlength{\aboverulesep}{0.5pt}
\setlength{\belowrulesep}{0.5pt}
\setlength\tabcolsep{0.8mm}
\setlength{\belowcaptionskip}{1.0mm}
\footnotesize
\centering
\caption{\textbf{Ablation study.} The default configuration, marked in \hlc{violet}, corresponds to the setting in \cref{tab:performance1}.}

\centering
(a) Ablation on the prompt composition.
\resizebox{\linewidth}{!}{
\begin{tabular}{
    >{\raggedright\arraybackslash}p{0.6cm}
    >{\centering\arraybackslash}p{1.8cm}
    >{\centering\arraybackslash}p{2.4cm}
    >{\centering\arraybackslash}p{2.4cm}|
    >{\centering\arraybackslash}p{1.8cm}
    >{\centering\arraybackslash}p{1.8cm}
    >{\centering\arraybackslash}p{1.8cm}
    >{\centering\arraybackslash}p{1.8cm}
    >{\centering\arraybackslash}p{1.8cm}
    >{\centering\arraybackslash}p{1.8cm}}
    \toprule
    \multirow{2}{*}{\textbf{No.}} &  \multicolumn{3}{c}{\textbf{Prompt composition}} & \multicolumn{6}{c}{\textbf{Metrics}}  \\
    \cmidrule(lr){2-4} \cmidrule(lr){5-10}
    & Criteria & Sequential tags & Reasoning order & Correctness & Specificity & Fluency & Relevance & Aggregation & Reversibility \\
    \midrule
    \textcolor{gray}{1} &  & \checkmark & forward &  $4.83\pm0.10$ & $4.58\pm0.14$ & $8.60\pm0.27$ & $5.32\pm0.16$ & $4.93\pm0.18$ & $0.58\pm0.00$  \\
    
    \textcolor{gray}{2} & \checkmark &  & forward & $5.27\pm0.06$ & $5.16\pm0.12$ & $8.63\pm0.28$ & $5.73\pm0.12$ & $5.35\pm0.09$ & $0.60\pm0.00$ \\
    
    \textcolor{gray}{3} & \checkmark & \checkmark & backward & $5.27\pm0.09$ & $4.94\pm0.16$ & $8.60\pm0.27$ & $5.81\pm0.11$ & $5.31\pm0.18$ & $0.62\pm0.00$ \\
    \rowcolor{violet!10}  \textcolor{gray}{4} & \checkmark & \checkmark & forward & $\mathbf{5.47\pm0.11}$ &  $\mathbf{5.16\pm0.18}$ & $\mathbf{8.65\pm0.30}$ & $\mathbf{5.98\pm0.13}$ & $\mathbf{5.51\pm0.20}$ & $\mathbf{0.62\pm0.00}$ \\
    \bottomrule
\end{tabular}}

\vspace{4pt}

\centering
(b) Ablation on the criteria composition.
\resizebox{\linewidth}{!}{
\begin{tabular}{
    >{\raggedright\arraybackslash}p{0.6cm}
    >{\centering\arraybackslash}p{2cm}
    >{\centering\arraybackslash}p{2.2cm}
    >{\centering\arraybackslash}p{2.4cm}|
    >{\centering\arraybackslash}p{1.8cm}
    >{\centering\arraybackslash}p{1.8cm}
    >{\centering\arraybackslash}p{1.8cm}
    >{\centering\arraybackslash}p{1.8cm}
    >{\centering\arraybackslash}p{1.8cm}
    >{\centering\arraybackslash}p{1.8cm}}
    \toprule
    \multirow{2}{*}{\textbf{No.}} &  \multicolumn{3}{c}{\textbf{Criteria composition}} & \multicolumn{6}{c}{\textbf{Metrics}}  \\
    \cmidrule(lr){2-4} \cmidrule(lr){5-10}
    & Ch. to Detect & Ch. to Ignore & Tone \& Nuance & Correctness & Specificity & Fluency & Relevance & Aggregation & Reversibility \\
    \midrule
    \textcolor{gray}{1} &  & \checkmark & \checkmark & $5.17\pm0.10$ & $4.91\pm0.15$ & $8.61\pm0.24$ & $5.63\pm0.16$ & $5.12\pm0.13$ & $0.60\pm0.00$ \\
    
    \textcolor{gray}{2} & \checkmark &  & \checkmark & $5.32\pm0.09$ & $5.05\pm0.14$ & $8.52\pm0.23$ & $5.85\pm0.16$ & $5.44\pm0.16$ & $0.61\pm0.00$ \\
    
    \textcolor{gray}{3} & \checkmark & \checkmark &  & $5.47\pm0.13$ & $5.16\pm0.19$ & $8.17\pm0.29$ & $5.98\pm0.16$ & $5.49\pm0.19$ & $0.60\pm0.00$ \\
    \rowcolor{violet!10}  \textcolor{gray}{4} & \checkmark & \checkmark & \checkmark & $\mathbf{5.47\pm0.11}$ &  $\mathbf{5.16\pm0.18}$ & $\mathbf{8.65\pm0.30}$ & $\mathbf{5.98\pm0.13}$ & $\mathbf{5.51\pm0.20}$ & $\mathbf{0.62\pm0.00}$ \\

    \bottomrule
\end{tabular}}
\label{ablation}
\end{table*}

\begin{figure}[t]
      \centering
      \includegraphics[width=\linewidth]{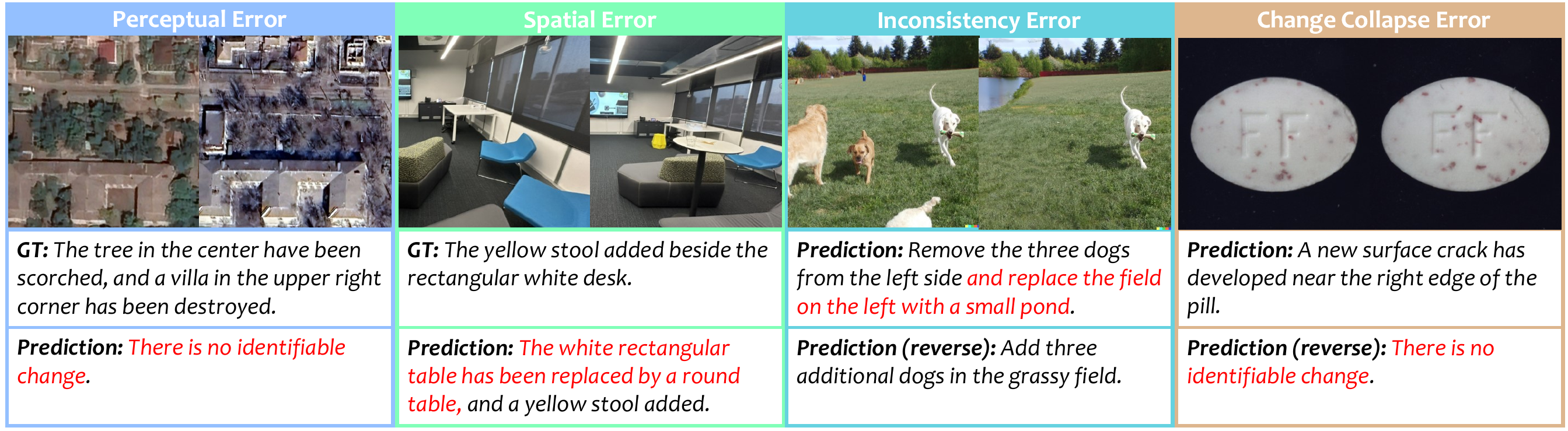}
      \vspace{-15pt}
    \caption{\textbf{Representative failure examples (Top-2 errors) of each error case in \textsc{C3-Bench}.} Erroneous regions are highlighted in \textcolor{red}{red}.}
  \label{fig:example}
\end{figure}

\noindent As shown in \cref{fig:error1}, perceptual error emerges as the primary bottleneck (63.3\% in total)---underscoring the significance of robust visual capabilities for accurate 
change captioning. The secondary bottleneck exhibits strong domain dependency, revealing distinct challenges across different domains. For instance, in the case of the Natural Scenes domain, where image pairs are often captured by moving vehicles or mobile robots \cite{Li2024UMADUO}, scene misalignment leads to a marked increase in spatial errors, causing models to confuse viewpoint-induced differences with genuine changes observed under a shared viewpoint. For reversibility-related errors, the inconsistency error is the most frequent type. However, within the Anomalies domain, change collapse errors dominate, where models fail to revert anomalous regions back to an intact, status quo state—even when the criteria explicitly require such recovery. This may suggest that the model’s implicit bias in anomalous image pairs predisposes it to focus on abnormal cues rather than restoring the status quo---highlighting that a misalignment between biases and the criteria can substantially affect the performance. \cref{fig:example} illustrates representative failure cases; more examples and analysis are provided in the Appendix.


\noindent\textbf{Why Models Cannot?} Reversibility errors may arise from models being uni-directionally trained on sequentially ordered image-text data and from the causal attention mechanism in their forward process. Description errors may stem from models being optimized for image-text alignment rather than for fine-grained perception or explicit geometry---lacking in visual and spatial understanding.

\subsection{Ablation Study on LMM Prompting}  
\label{sec:ablation}

To gain further insight into LMM-based change captioning, we ablate key prompt components---change criteria, sequential tags, and reasoning order (\textit{before}-image-first vs. \textit{after}-image-first). As illustrated in \cref{ablation} (a), removing explicit criteria and relying on generic instructions leads to substantial degradation, indicating that model’s implicit biases are misaligned with context-relevant semantics---underlining the importance of our criteria-conditioned prompt design over previous naive prompting schemes \cite{Zhong2025DECIDERDC, Hu2025MCTCCDiffCC, anonymous2026imagine, Zhang2024DifferentialPerceptiveAR}; while all components contribute meaningfully, Change to Detect plays a key role in enhancing contextual alignment (see \cref{ablation}(b)). Excluding image tags also deteriorates performance, underscoring the advantages of explicitly grounding sequential order for accurate and reversible change descriptions. Finally, backward reasoning, presenting the \textit{after}-image first, further impairs performance, indicating that misalignment between image order and reasoning direction induces additional position bias. This finding suggests that LMMs are biased 
to canonical temporal ordering (\textit{before}-first), likely due to training-time exposure to temporally ordered image-text data.

\begin{table*}[t]
\renewcommand{\arraystretch}{1.0}
\setlength{\aboverulesep}{0.5pt}
\setlength{\belowrulesep}{0.5pt}
\setlength\tabcolsep{0.8mm}
\setlength{\belowcaptionskip}{1.0mm}
\footnotesize
\centering
\caption{\textbf{Results of multi-dataset training.} We use CLIP4IDC \cite{Guo2022CLIP4IDCCF}, the strongest conventional baseline. See \cref{protocol} for datasets included in ALL.}

\centering
\resizebox{\linewidth}{!}{
\begin{tabular}{
    >{\raggedright\arraybackslash}p{0.6cm}
    >{\centering\arraybackslash}p{4.3cm}
    >{\centering\arraybackslash}p{2.1cm}
    >{\centering\arraybackslash}p{2.1cm}
    >{\centering\arraybackslash}p{2.1cm}
    >{\centering\arraybackslash}p{2.1cm}
    >{\centering\arraybackslash}p{2.1cm}}
    \toprule
    \multirow{2}{*}{\textbf{No.}} & \multirow{2}{*}{\textbf{Training}} & \multicolumn{5}{c}{\textbf{Metrics}}  \\
    \cmidrule(lr){3-7}
    &  & Correctness & Specificity & Fluency & Relevance & Aggregation \\
    \midrule
    \textcolor{gray}{1} & Single (LEVIR-CC \cite{Liu2022RemoteSI}) & $1.78\pm0.02$ & $1.63\pm0.03$ & $8.81\pm0.23$ & $2.35\pm0.22$ & $1.79\pm0.03$ \\
    
    \rowcolor{LGray}\textcolor{gray}{2} & ALL \cite{Park2019RobustCC, Liu2022RemoteSI, Jhamtani2018LearningTD, Tan2019ExpressingVR} & $1.31\pm0.01$ & $1.24\pm0.02$ & $4.21\pm0.34$ & $1.58\pm0.09$ & $1.31\pm0.02$ \\
    \bottomrule
\end{tabular}}
\label{ablation2}
\vspace{-3pt}
\end{table*}


\subsection{Multi-Dataset Training for Change Captioning}
Can training on multiple datasets concurrently improve change captioning performance, as in other data-driven computer vision tasks \cite{Park2022DualTL, Yang2024DepthAU, Yang2024DepthAV2, Jeong2025TestTimePT}? As shown in \cref{ablation2}, simple unification does not lead to a performance gain---even worsening the results. This can be attributed to two main factors: \textit{(i)} the size of the unified dataset is still insufficient to induce a discriminative latent space to bridge the semantic gap across heterogeneous change contexts, and \textit{(ii)} contextual conflicts between the implicit criteria of each dataset (e.g., color differences are treated as meaningful in CLEVR-CC \cite{Park2019RobustCC}, whereas they are regarded as pseudo-changes in LEVIR-CC \cite{Liu2022RemoteSI}; and IER \cite{Tan2019ExpressingVR} requires an instructional tone whereas most other datasets adopt a descriptive tone) confuse the overall learning signal, highlighting the significance of our context-aware formulation for coherent data scaling. 


\section{Conclusion}
In this paper, we introduced \textsc{C3-Bench}, a comprehensive benchmark for evaluating Context-aware Change Captioning. Based on a distinct problem formulation, \textsc{C3-Bench} provides a dataset with unprecedented diversity spanning multiple domains and contexts, along with principled metrics, establishing a robust testbed for both conventional models and modern LMMs. Through extensive experiments, we revealed key challenges of existing benchmarks and model families, including the limited generalizability of conventional models, critical bottlenecks in LMMs, pronounced pairwise position bias, and crucial components in prompt design. We expect \textsc{C3-Bench} to lay a solid foundation and guide future research toward more robust and reliable change understanding.

\section*{Acknowledgements}

This research was partly supported by Institute of Information \& communications Technology Planning \& Evaluation (IITP) grant funded by the Korea government (MSIT) (No. RS-2022-II220926, Development of Self-directed Visual Intelligence Technology Based on Problem Hypothesis and Self-supervised Methods); by the InnoCORE program of the Ministry of Science and ICT (GIST InnoCORE KH0860); by the National Research Foundation of Korea (NRF) grant funded by the Korea government (MSIT) (No. NRF-2022R1C1C1009989).

\bibliographystyle{splncs04}
\bibliography{main}
\end{document}